\documentclass[a4paper,fleqn]{cas-dc}

\usepackage[numbers]{natbib}
\usepackage{multirow}
\usepackage{multicol}
\usepackage{csquotes}
\usepackage{amsmath}
\usepackage{amssymb}
\usepackage{algorithm}
\usepackage{algpseudocode}
\usepackage{graphicx}
\usepackage{subcaption}
\usepackage{footnote}
\usepackage{soul}
\usepackage{hyperref}
\def\tsc#1{\csdef{#1}{\textsc{\lowercase{#1}}\xspace}}
\tsc{WGM}
\tsc{QE}
\tsc{EP}
\tsc{PMS}
\tsc{BEC}
\tsc{DE}

\begin{document}
\let\WriteBookmarks\relax
\def\floatpagepagefraction{1}
\def\textpagefraction{.001}

\shorttitle{Fine-grained Approach to Pediatric Wrist Pathology Recognition on a Limited Dataset}

\title [mode = title]{Learning from the Few: Fine-grained Approach to Pediatric Wrist Pathology Recognition on a Limited Dataset}                      
\tnotetext[1]{This work was supported in part by the Department of Computer Science (IDI), Faculty of Information Technology and Electrical Engineering, Norwegian University of Science and Technology (NTNU), Gjøvik, Norway; and in part by the Curricula Development and Capacity Building in Applied Computer Science for Pakistani Higher Education Institutions (CONNECT), Project number: NORPART-2021/10502, funded by DIKU.}

%
\author[1]{Ammar Ahmed}[type=editor,
                        auid=000,bioid=1]


\ead{@ammaa@stud.ntnu.no}


\credit{Conceptualization, methodology, data collection, original draft preparation, writing}

\affiliation[1]{organization={Intelligent Systems and Analytics (ISA) Research Group, Department of Computer Science (IDI), Norwegian University of Science \& Technology (NTNU)},
    city={Gj\o vik},
    postcode={2815}, 
    country={Norway}}

\author[1]{Ali Shariq Imran}
\cormark[1]
\ead{ali.imran@ntnu.no}
\credit{Methodology, review and editing, supervision}

\author%
[2]
{Zenun Kastrati}
\ead{zenun.kastrati@lnu.se}
\credit{Methodology, review and editing, supervision}

\affiliation[2]{organization={Department of Informatics, Linnaeus University}, 
    city={Växjö},
    postcode={351 95}, 
    country={Sweden}}

\author%
[3]
{Sher Muhammad Daudpota}
\ead{sher@iba-suk.edu.pk}
\credit{Methodology, review and editing, supervision}

\affiliation[3]{organization={Department of Computer Science, Sukkur IBA University}, 
    city={Sukkur},
    postcode={65200}, 
    country={Pakistan}}

\author%
[1]
{Mohib Ullah}
\ead{mohib.ullah@ntnu.no}
\credit{Review and editing, supervision}
\author%
[4]
{Waheed Noor}
\ead{waheed.noor@um.uob.edu.pk}
\credit{Methodology, review and supervision}

\affiliation[4]{organization={Department of Computer Science \& Information Technology,, University of Balochistan}, 
    city={Quetta},
    postcode={87300}, 
    country={Pakistan}}

\cortext[cor1]{Corresponding author}

\begin{abstract}
Wrist pathologies, {particularly fractures common among children and adolescents}, present a critical diagnostic challenge. While X-ray imaging remains a prevalent diagnostic tool, the increasing misinterpretation rates highlight the need for more accurate analysis, especially considering the lack of specialized training among many surgeons and physicians. Recent advancements in deep convolutional neural networks offer promise in automating pathology detection in trauma X-rays. However, distinguishing subtle variations between {pediatric} wrist pathologies in X-rays remains challenging. Traditional manual annotation, though effective, is laborious, costly, and requires specialized expertise. {In this paper, we address the challenge of pediatric wrist pathology recognition with a fine-grained approach, aimed at automatically identifying discriminative regions in X-rays without manual intervention. We refine our fine-grained architecture through ablation analysis and the integration of LION.} Leveraging Grad-CAM, an explainable AI technique, we highlight these regions. Despite using limited data, reflective of real-world medical study constraints, our method consistently outperforms state-of-the-art image recognition models on both augmented and original (challenging) test sets. {Our proposed refined architecture achieves an increase in accuracy of 1.06\% and 1.25\% compared to the baseline method, resulting in accuracies of 86\% and 84\%, respectively. Moreover, our approach demonstrates the highest fracture sensitivity of 97\%, highlighting its potential to enhance wrist pathology recognition.} The implementation code can be found at https://github.com/ammarlodhi255/fine-grained-approach-to-wrist-pathology-recognition
\end{abstract}




\maketitle

\section{Introduction}\label{sec1}
Wrist pathologies, whether congenital (present at birth) or acquired (developed later in life), are major concerns. During puberty, there's a marked increase in the incidence of wrist fractures, particularly of the distal radius and ulna, among children, adolescents, and young adults \cite{hedstrom_2010, randsborg_et_al_2013}. Annually, tens of millions globally experience hand injuries \cite{Er2013}. Diagnosis is primarily conducted using X-ray imaging, a method that continues to be widely used \cite{RSNA_ACR2023} even though it was developed over a century ago. {X-rays} procured are subsequently evaluated by {surgeons, physicians, or radiologists} to identify {pediatric} wrist pathologies. Nonetheless, many healthcare practitioners might not possess the specialized training needed for accurate injury assessment, potentially leading them to interpret {X-rays} without the guidance of seasoned radiologists or skilled peers \cite{Hallas_Ellingsen_2006}. Research indicates that misdiagnoses from interpreting emergency X-rays can be as high as 26\% \cite{Guly_2001, Mounts_2011}. 

Recently, deep convolutional neural networks (CNNs) have demonstrated noteworthy efficacy in detecting pathologies in trauma X-rays \cite{adams2020ai, tanzi2020fracture, choi2020dual, ahmed2024enhancing, Ahmed2024}. For optimal performance, CNNs ideally require training on a vast number of samples, given the intrinsic complexity of image recognition—a task that involves learning intricate functions for category discrimination. This challenge amplifies when the target categories possess similar visual characteristics \cite{DSSE}. Nonetheless, procuring extensive X-ray datasets is fraught with difficulties, making the acquisition of millions of samples for classification often impractical. Therefore, in many cases, researchers have access to very limited medical data, due to it being laborious and very time-consuming to acquire \cite{Chang2023, Mehta2023}.

Distinguishing between wrist pathologies in X-rays poses a challenge due to their subtle variations. One approach could be manual annotation to emphasize the discriminative regions showcasing these pathologies. However, this method is labor-intensive, expensive, and necessitates specialized domain knowledge \cite{kmlvision2023}. The pressing question remains: how can we differentiate between similar-looking categories without the manual annotation and extensive data typically required for such intricate recognition tasks?

In the realm of computer vision, when confronted with limited data, a common approach is \enquote{hand-engineering}. This encompasses tasks from meticulous manual annotation to crafting specialized components within CNN architectures \cite{Dhami2018}, ensuring they are optimally designed for specific recognition challenges such as designing an architecture that purposefully looks for the most discriminative regions and figures out patterns even from small data. Since manual annotation is infeasible due to challenges highlighted earlier, we, therefore, require a specialized architecture for discrimination of categories with highly similar appearance. In this study, we approach wrist pathology recognition as a \enquote{fine-grained recognition} problem.

Existing fine-grained architectures fall into three categories: (1) RPN-based region identification \cite{Ren2017}, (2) feature map amplification through attention mechanisms \cite{Rao2021, Zheng2017, Sun2018, Zhuang2020, Hu2019}, and (3) self-attention-driven region discrimination \cite{He2021, Wang2021}. The first two methods use CNNs with complex, multi-stage architectures that hinder end-to-end training. Based on vision transformers (ViT), the third method utilizes self-attention for region selection but lacks hierarchical feature representation for local details \cite{Chou2022}. Additionally, these architectures often produce large prediction regions, making it challenging to capture subtle local features.


In this study, we utilize the recent Plug-in Module framework for fine-grained visual recognition that can be trained end-to-end \cite{Chou2022}. This method uses novel background segmentation, pixel-level predictions, and feature fusion techniques, which can effectively improve the accuracy of fine-grained visual recognition of wrist pathologies. The fundamental design principle is to consider each pixel on a feature map as a distinct feature, symbolizing its unique region, as many of these pathologies such as fractures can be very small in size \cite{Yahalomi2018}. We determine the significance of each pixel based on its predicted class score and subsequently merge the critical pixels. We later integrate \textit{Evo\textbf{l}ved S\textbf{i}gn M\textbf{o}me\textbf{n}tum} (LION) optimizer \cite{Chen2023} recently released to the public, to improve the network's generalization capabilities as well as its overall performance. Additionally, we conducted an ablation study to investigate various components of the network. This comprehensive analysis concludes with the proposal of a tailored network optimized for the specific task of wrist pathology recognition in a fine-grained manner on a limited dataset. 
To the best of our knowledge, this is the first time a fine-grained analysis has been done on real-world X-rays involving different projections and the presence of several pathologies. We hypothesize that our employed fine-grained approach will outperform many of the well-known and state-of-the-art CNN techniques.


This paper presents the following primary contributions:
\begin{itemize}
\item Reframing wrist pathology recognition as a fine-grained recognition problem.
\item Utilizing a fine-grained recognition framework on a carefully curated, limited dataset showcasing superior performance compared to existing recognition techniques.
\item Introducing a refined fine-grained architecture for the task of wrist pathology recognition subsequent to LION optimizer integration and component adjustment.
\item Integration of XAI in fine-grained architecture to pinpoint discriminative regions in wrist X-rays.
\end{itemize}

 The rest of the article is structured as follows. Section \ref{sec2} provides an overview of the fine-grained visual recognition. Section \ref{sec3} provides an overview of related research. In Section \ref{sec4}, we detail the methodology employed in our study. Section \ref{sec5} presents the results and facilitates a thorough discussion. Lastly, Section \ref{sec6} concludes this paper with potential future research directions.

\section{Fine-grained Visual Recognition (FGVR)}\label{sec2}
 
 Visual recognition can be categorized into coarse-grained and fine-grained recognition. Coarse-grained recognition deals with categorizing highly dissimilar-looking classes such as differentiating between cats and dogs. In contrast, fine-grained recognition involves identifying categories with subtle differences, such as distinguishing bird or dog species. More formally, recognition categories can be seen as tasks of categorizing elements into classes with different levels of granularity. Let $\mathcal{X}$ be the space of data instances and $\mathcal{Y}_C$ and $\mathcal{Y}_F$ be the set of classes in coarse-grained and fine-grained tasks, respectively. The coarse-grained recognition task is to learn a function $f_C: \mathcal{X} \rightarrow \mathcal{Y}_C$ such that the dissimilarity between categories is maximized. This can be formalized as:
\begin{equation}
\arg\max_{f_C} \sum_{\substack{i, j \ f_C(x_i) \neq f_C(x_j)}} D\left(f_C(x_i), f_C(x_j)\right),
\end{equation}
where, $D$ is a dissimilarity measure, and $x_i, x_j \in \mathcal{X}$ are data instances from different classes. On the other hand, the fine-grained recognition task is to learn a function $f_F: \mathcal{X} \rightarrow \mathcal{Y}_F$ such that the similarity within the same category is maximized and the dissimilarity between different subcategories is maximized. This can be formalized as:
\begin{align}
    \arg\max_{f_F} & \sum_{i, j: f_F(x_i) = f_F(x_j)} S(x_i, x_j) \\
    &- \alpha \sum_{\substack{i, j \\ f_F(x_i) \neq f_F(x_j)}} D(x_i, x_j),
\end{align}
where $S$ is a similarity measure, $\alpha$ is a trade-off parameter, and $x_i, x_j \in \mathcal{X}$ are data instances.

\begin{figure}[!h]
\centering
\includegraphics[width=0.9\linewidth, height=4.5cm]{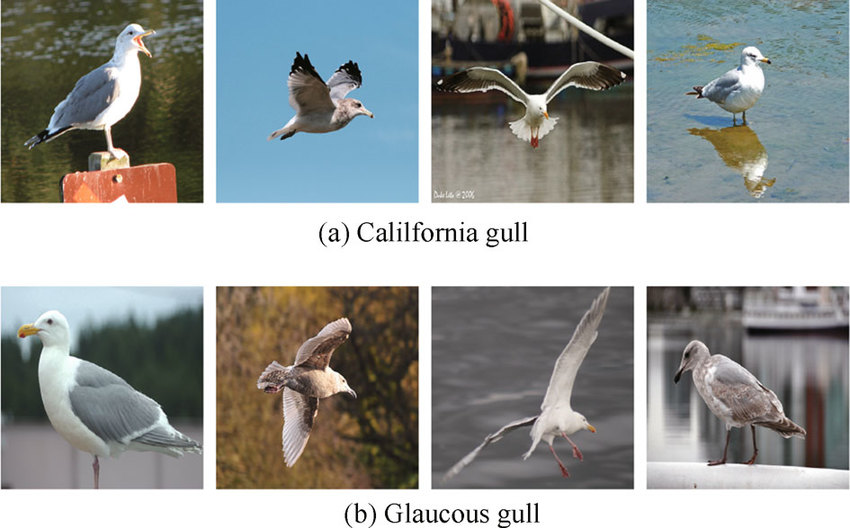}
\caption{Two species of gulls from the CUB200 dataset, illustrating the difficulty of fine-grained recognition.}
\label{fig1:birds}
\end{figure}

The intricacies of fine-grained visual recognition present three primary challenges \cite{Chou2022, zhao1}. Firstly, there is a large intra-class variance such that a single category can exhibit significant variation. For instance, images of a single bird type could be captured from various angles and poses as illustrated in Fig.\ref{fig1:birds}. Secondly, there is a small inter-class variance; objects from distinct subcategories might bear a striking resemblance to one another. The similarities of the two birds in Fig.\ref{fig1:birds} serve as an example. Lastly, unlike coarse-grained recognition, fine-grained categorization typically mandates expert professionals for data labeling, thereby elevating the costs associated with data collection.

\begin{figure}[!h]
\centering
\includegraphics[width=1\linewidth, height=6cm]{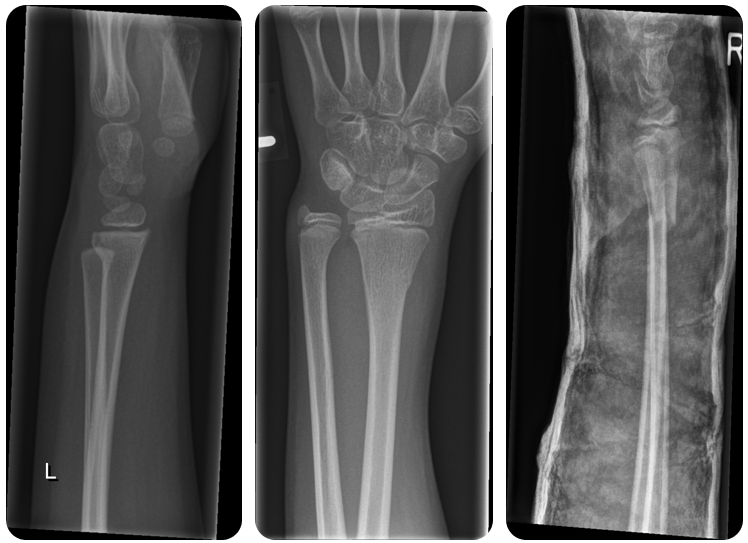}
\caption{Large intra-class variance within fracture pathology.}
\label{fig1:wrist_intra}
\end{figure}

\begin{figure}[!h]
\centering
\includegraphics[width=1\linewidth, height=6cm]{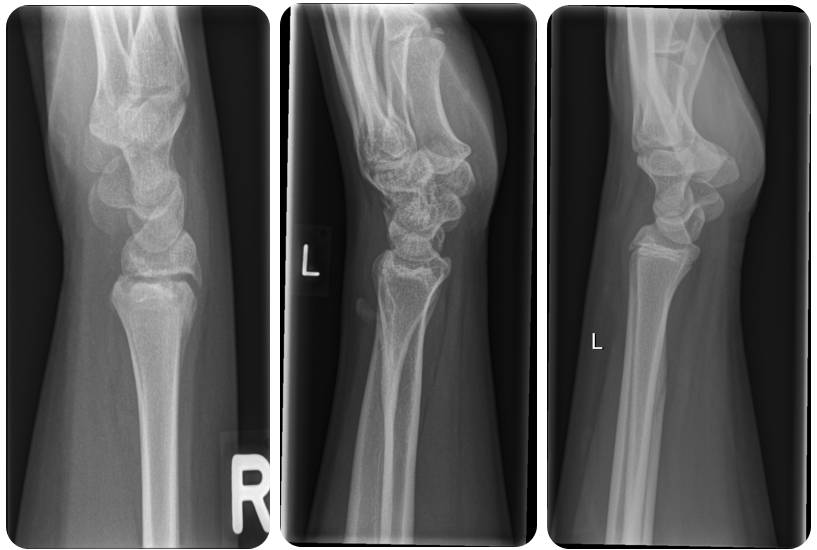}
\caption{Small inter-class variance between three different pathologies: fracture (left), bone anomaly (middle), and soft tissue (right), respectively.}
\label{fig1:wrist_inter}
\end{figure}

Adapting the primary challenges of fine-grained recognition to our domain of wrist pathology identification, we recognize the following intricacies. First, there exists pronounced intra-class variability. For instance, consider three distinct wrist X-rays depicting fractures, showcased in Fig.\ref{fig1:wrist_intra}. Here, the fracture pathology is captured in various angles, including posteroanterior and lateral views. Second, the data reveals subtle inter-class differences among the pathologies. As an illustration, Fig.\ref{fig1:wrist_inter} displays three images, each representing a unique pathology: fracture, bone anomaly, and soft tissue, respectively. It can be seen that it is extremely hard to distinguish between these categories. The third pivotal challenge is that fine-grained recognition typically demands expertise, necessitating a radiologist in our context to spotlight the distinguishing regions manually. While one could bear the expenses and time commitments of manual annotations, we are confronted with another predicament: a significant shortage of radiologists, even in developed nations \cite{Makary_Takacs_2022, Burki_2018, Rimmer_2017}. Not to mention, this shortage is expected to increase in coming years \cite{marlow_et_al_2019}. Thus, depending on 
manual annotation for the long term becomes impractical.

\section{Related Work}\label{sec3}
While our study emphasizes wrist pathology recognition, including fractures, recent research has predominantly targeted fracture detection. This section presents a review of wrist fracture detection and classification studies, highlighting their key findings and specifically covering advancements from the past five years. Additionally, we also discuss the most recent applications of fine-grained classification in other medical domains. {To conclude this section, we explore XAI approaches in various medical domains and an alternative approach of self-supervised learning tailored for datasets with limited availability.}

Thian et al \cite{thian2019convolutional}, trained Faster R-CNN models on 6,515 frontal and 6,537 lateral wrist projections. The frontal model achieved a specificity of 0.83 and a sensitivity of 0.96, while the lateral achieved 0.86 and 0.9, respectively. Both had an overall specificity of 0.73, a sensitivity of 0.98, and AUC-ROC of 0.89. Hržić et al. \cite{hrzic2019}, improved bone visibility in wrist X-rays using local entropy, followed by fracture detection. A graph-based technique enhanced bone contours and edge detection. They achieved 0.91 accuracy and 0.86 sensitivity on a dataset of 860 child radius and ulna bone X-ray images (642 fracture-free, 218 with fractures).

Guan et al. \cite{Guan2020} employed a two-stage R-CNN technique and secured an average precision (AP) of 0.62 from roughly 4,000 arm fracture X-ray images in the MURA dataset. In a similar vein, Wang et al. \cite{wang_yao_zhang_guan_wang_2021} introduced ParallelNet, a two-stage R-CNN structure underpinned by the TripleNet backbone, to detect fractures. They tested this on a collection of 3,842 thigh fracture X-ray images and registered an AP of 0.88. Kandel et al. \cite{kandel2020musculoskeletal}, fine-tuned six CNNs for musculoskeletal image classification. Transfer learning outperformed training from scratch. The top performer for wrist images was the fine-tuned DenseNet121 with fully connected layers, achieving a mean accuracy of 0.82.

Raisuddin et al. \cite{raisuddin2021} developed DeepWrist, a deep-learning pipeline for distal radius fracture detection. Trained on 3873 images, the model achieved AP scores of 0.99 and 0.64 on two test sets—207 cases and 105 challenging cases, respectively. It generated heatmaps showing fracture likelihood in the wrist area but lacked precise fracture localization. The study was limited by a small dataset and an imbalanced distribution of challenging cases. Xue et al. \cite{Xue2021} introduced a guided anchoring approach to detect fractures in hand X-ray images with Faster R-CNN. The evaluation, using 3067 images, yielded an AP score of 0.71. Ma et al. \cite{ma2021bone} initially categorized Radiopaedia dataset images into fracture and non-fracture using CrackNet. Subsequently, they employed Faster R-CNN to detect fractures in the 1052 bone images, achieving an accuracy, recall, and precision of 0.88, 0.88, and 0.89, respectively. Kim et al. \cite{kim2021} gathered image data from patients presenting wrist trauma at the emergency department. They utilized two CNN models, namely DenseNet-161 and ResNet-152. To identify influential regions in {X-ray} scans, gradient-weighted class activation mapping was employed. DenseNet-161 demonstrated a sensitivity and accuracy of 0.90. For ResNet-152, both values were 0.89. The AU-ROC for wrist fracture detection was 0.96 for DenseNet-161 and 0.95 for ResNet-152.

Joshi et al. \cite{Joshi2022} used transfer learning with a modified Mask R-CNN on two datasets: 3000 surface crack images and 315 wrist fracture images. After training on the crack dataset and fine-tuning on the wrist dataset, they achieved 0.92 AP for detection and 0.78 for segmentation. Hardalac et al. \cite{hardalac2022} performed 20 fracture detection experiments using wrist X-ray images from Gazi University Hospital. Their ensemble model, WFD-C, combining five models, achieved the highest AP of 0.86. Hržić et al. \cite{hrzic2022fracture} compared YOLOv4 and U-Net for fracture detection on the \enquote{GRAZPEDWRI-DX} dataset. YOLOv4 outperformed with higher AUC-ROC (up to 0.90) and F1 scores (up to 0.96) for fracture detection.  Ahmed et al. \cite{ahmed2024enhancing} demonstrated the effectiveness of single-stage models in improving pediatric wrist imaging. By comparing compound-scaled versions of different YOLO models, where YOLOv8m exhibited the highest fracture detection sensitivity of 0.92 and mAP of 0.95. Conversely, YOLOv6m achieved the highest sensitivity across all classes, reaching 0.83. Additionally, YOLOv8x achieved the highest mAP of 0.77 across all classes in the GRAZPEDWRI-DX pediatric wrist dataset.

There is a bunch of papers that discuss the most recent applications of fine-grained classification in other medical domains. For instance, Kumar et al. \cite{Kumar2023} developed a CNN model specifically designed for fine-grained skin cancer classification. Their study revealed the significance of incorporating features from early, intermediate, and final layers to enhance the accuracy and robustness of the model. The proposed model achieved a peak accuracy of 74\%. Lu et al. \cite{Lu2023} used a method comprising a localization and a fusion module aimed at addressing the challenges of fine-grained classification due to the varied shape and size of lesions and the intricate relationship between lesions and the background. Their approach involved identifying and extracting multiple lesions of various shapes and sizes from the original image. Further, they fused the features of lesions and backgrounds using an attention mechanism. Evaluating their model on two clinical datasets yielded accuracies of 80.72\% and 84.68\%, respectively. Park et al. \cite{Park2023} proposed a Fine-Grained Self-Supervised Learning (FG-SSL) technique, which employs a hierarchical block to facilitate gradual learning. This method aims to minimize the cross-correlation between fine-grained Jigsaw puzzle images and regularized original images, approximating an identity matrix. The approach was evaluated on popular datasets including ISIC2018, APTOS2019, and ISIC2017, yielding an accuracy of 85.8\%, 89.7\%, and 91.3\%, respectively. Xiao et al. \cite{Xiao2019} focused on classifying endoscopic tympanic membrane images using the FGVC approach. They incorporated a localization technique in their process. Experimental results demonstrated that selecting patches based on heatmap-guided localization outperformed random cropping. Their pipeline, utilizing a ResNet backbone, achieved the highest accuracy of 94\%. Liu et al. \cite{Liu2020} introduced a deep learning approach, leveraging bilinear convolutional neural networks (BCNNs), to perform fine-grained classification of breast cancer histopathological images. Employing bilinear pooling, they aggregated numerous orderless features without considering the disease's location. Their experiments conducted on the BreaKHis breast cancer dataset yielded a 95.95\% accuracy in fine-grained classification. Fan et al. \cite{Fan2020} Fan et al. introduced an attention-based deep architecture aimed at enhancing background suppression and recognizing crucial instances within images. Their method focused on automatically localizing discriminative instances in fine-grained images without additional supervision or redundant region proposals. Their approach demonstrated an accuracy of 94.3\% on a benchmark dataset for Breast Cancer Histology.

{Recently, Grad-CAM} \cite{selvaraju2019gradcam}, {has gained popularity as an explainability tool in computer vision tasks. Its appeal lies in its simplicity and effectiveness in highlighting regions of an image that contribute most significantly to a model's prediction.
While Grad-CAM is a popular choice by researchers in various medical domains} \cite{ali2023enlightening,ali2023visualizing}, {it's essential to acknowledge other XAI approaches and consider their merits. Some of the other well-known XAI techniques include LIME} \cite{ribeiro2016why} {and SHAP} \cite{lundberg2017unified}, 
{Both of which fall under the category of Perturbation-based methods, CAM} \cite{zhou2016learning}, 
{and counterfactual explanations also termed ``what-if'' explanations.
Gulum et al.} \cite{gulum2021improved} {combines two saliency methods, namely Saliency Maps and Grad-CAM to improve the robustness and accuracy of localization of the prostate lesion. Lee et al. introduced Pyramid-CAM} \cite{lee2018robust} {for brain tumor localization, while Shen et al.} \cite{shen2019globally} {employed Saliency Maps for breast lesion localization. Arun et al.} \cite{arun2020assessing} {illustrate that Grad-CAM surpasses Saliency Maps in cascading randomization experiments. Corizzo et al.} \cite{9671335} {leveraged deep neural networks to extract a representation of images and further analyze them through clustering, dimensionality reduction for visualization, and class activation mapping. Ramy et al.} \cite{zeineldin2022explainability} {proposed NeuroXAI, which incorporated seven explanation techniques, offering visualization maps to enhance the interpretability of deep learning models. The framework was utilized for image classification and segmentation tasks using magnetic resonance (MR) imaging modality. Hurtado et al.} \cite{hurtado2022bioinformatics} {empirically and technically explore the interpretability of contemporary XAI techniques, specifically LIME and SHAP, focusing on result reproducibility and execution time using a melanoma image classification dataset. LIME performed better than SHAP gradient explainer in terms of reproducibility and execution time. The paper by Wachter et al.} \cite{wachter2017counterfactual} {discusses the significance of counterfactual explanations in automated decision-making systems. Mertes et al.} \cite{mertes2022ganterfactual} {proposed GANterfactual, an approach to generate such counterfactual image explanations based on adversarial image-to-image translation techniques and evaluated their approach in an exemplary medical use case. More recently, Bedel et al.} \cite{bedel2023dreamr} {introduced a Diffusion-driven Counterfactual Explanation, a novel explanation method for funtional MRI, which they termed as ``DreaMR''. It employs diffusion-based resampling of fMRI data to manipulate downstream classifier decisions, subsequently determining the minimal deviation between original and counterfactual samples for explanatory purposes. Unlike traditional diffusion techniques, DreaMR integrates a fractional multi-phase-distilled diffusion prior, enhancing sampling efficiency while maintaining fidelity. Additionally, it utilizes a transformer architecture to incorporate long-range spatiotemporal context within fMRI scans. Experimental validations reveal DreaMR's enhanced efficacy compared to alternative counterfactual techniques utilizing VAE, GAN, and conventional diffusion priors. Notably, DreaMR's capability to generate differences between original and counterfactual samples facilitates more precise and credible explanations than baseline methods.} 

{Given the exceedingly restricted dataset utilized in this study, we cannot leave out the discussion on the effectiveness of self-supervised and few other methodologies in such circumstances, aside from fine-grained recognition techniques employed in this study. Self-supervised learning can be particularly useful in scenarios where obtaining labeled data is expensive or impractical. By leveraging unlabeled data, self-supervised methods can still learn meaningful representations, which can then be fine-tuned on a small amount of labeled data for specific tasks, thus maximizing the utility of limited data resources. Several studies have demonstrated the effectiveness of the self-supervised learning throughout various medical image analysis tasks } \cite{lu2020semi, li2021rotation, sriram2021covid, chen2019self, sowrirajan2021moco, nguyen2020self, taleb2020self, xie2020pgl, chaitanya2020contrastive}. {Recently, Korkmaz et al.} \cite{korkmaz2023} {introduced ``SSDiffRecon'' a self-supervised deep reconstruction model designed to mitigate various challenges associated with image fidelity, contextual sensitivity, and dependence on fully-sampled acquisitions for model training. SSDiffRecon embodies a conditional diffusion process as an unrolled architecture, integrating cross-attention transformers for reverse diffusion steps alongside data-consistency blocks for physics-driven processing. Unlike recent MRI diffusion techniques, SSDiffRecon employs a self-supervision approach, training solely on undersampled k-space data. Extensive experiments conducted on publicly available brain MR datasets ``fastMRI'' and ``IXI'' demonstrated the superior performance of SSDiffRecon over state-of-the-art supervised and self-supervised benchmarks in terms of both reconstruction speed and quality. Dar et al. } \cite{dar2023parallel} {introduced PSFNet, a parallel-stream fusion model for MRI reconstruction, merging SS and SG priors to enhance performance in low-data regimes while maintaining competitive inference times. Their approach combined nonlinear SG priors with linear SS priors via a parallel-stream architecture with learnable fusion parameters to alleviate error propagation under low-data regimes. Demonstrations on multi-coil brain MRI show PSFNet outperforms SG methods in low-data scenarios and surpasses SS methods with few training samples achieving, on average, 3.1 dB higher PSNR, 2.8\% higher SSIM, and 0.3 times lower RMSE than baselines. Öztürk et al.} \cite{ozturk2022content} {proposed a triplet-learning method for automated querying of medical image repositories using a Opponent Class Adaptive Margin (OCAM) loss. OCAM adjusts the margin value continually during training to maintain optimally discriminative representations. OCAM's performance is compared against state-of-the-art loss functions on three public databases such as gastrointestinal disease, skin lesion, and lung disease, showing superior performance in each application domain. 
Rehman et al.} \cite{rehman2024hybrid} {developed a CKD prediction model using logistic regression, LDA, and MLP classifiers based on MRI features. Logistic regression achieved accuracies of 98.5\% (train) and 97.5\% (test), while LDA and MLP attained 96.07\% and 95\% accuracy, respectively, for the train dataset. Serum creatinine, albumin, diabetes mellitus, red blood cell count, pus cell, and hypertension emerged as key discriminating features for CKD.}


It is evident from the recent fracture detection and classification literature that most studies (with a few exceptions) employ object detection on a small, manually annotated X-ray dataset. {Moreover, while self-supervised learning is a viable option for scenarios with limited data, fine-grained recognition holds particular advantages for wrist pathology X-rays. It can excel in this context due to its ability to discern subtle and nuanced differences within specific classes or categories of pathology, such as identifying variations in bone structure, joint alignment, or tissue abnormalities unique to wrist pathologies. This level of granularity is essential in accurately diagnosing and distinguishing between various wrist conditions, which often present with subtle differences that may be challenging to detect using traditional methods.} In our work, we take a novel fine-grained recognition approach to identify wrist pathologies, including fractures. By targeting discriminative X-ray regions and employing Grad-CAM, we eliminate the need for manual bounding boxes. 

 
\section{Methodology}\label{sec4}
\begin{figure*}
\centering
\includegraphics[width=1\textwidth, height=5.0cm]{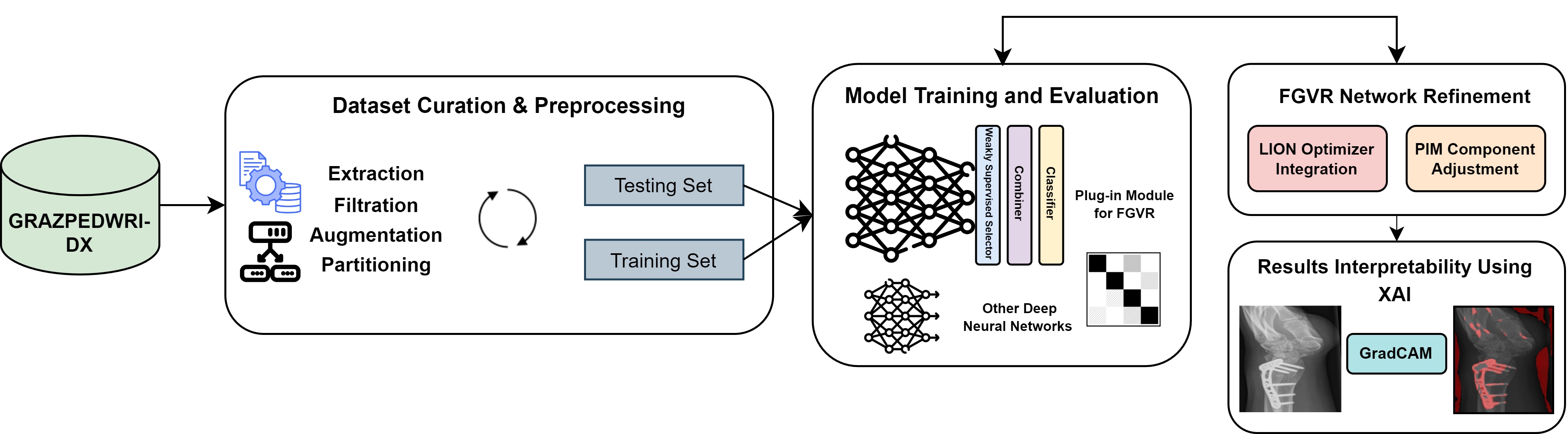}
\caption{Proposed methodology depicting dataset curation, model training and evaluation, network refinement, and interpretability.}
\label{fig1:methodology}
\end{figure*}

In this section, we outline our methodology, beginning with the steps taken in dataset curation such as extraction, filtration, augmentation, and partitioning. Subsequently, we delve into the components and operational principles of the Plug-in Module for FGVR. Following this, we explore the LION optimization algorithm. Lastly, we offer insights into our experimental settings, including the utilization of other deep learning models and training specifics. Figure \ref{fig1:methodology} offers a visual overview of our methodology.

\subsection{Dataset Curation}
We tailored a custom dataset, adapting it from the recently released \enquote{GRAZPEDWRI} dataset, designed initially for object detection purposes \cite{nagy2022pediatric}. The dataset is characterized by its focus on pediatric wrist {X-rays}, featuring images in PNG format from 6,091 patients with a mean age of 10.9 years (ranging from 0.2 to 19 years) meticulously collected by the Division of Paediatric Radiology within the Department of Radiology at the Medical University of Graz, Austria. The dataset includes 2,688 female patients, 3,402 male patients, and 1 with an unknown gender. It consists of a total of 20,327 wrist images, encompassing both lateral and posteroanterior projections. It includes nine distinct objects. The \enquote{text} object serves the purpose of identifying the hand side.



The dataset presents a challenge in that numerous instances contain multiple objects within a single image (it can contain a fracture as well as other pathologies). Given our study's emphasis on multi-class recognition, as opposed to multi-label recognition, our approach requires extracting images from the dataset that exclusively represent a single object/class per image. 
The dataset also presents a second challenge in the form of significant class imbalance, with most instances belonging to the \enquote{fracture} class. Furthermore, this imbalance will be exacerbated after the extraction of single-class images.

\begin{figure}
\includegraphics[width=1\linewidth, height=5.5cm]{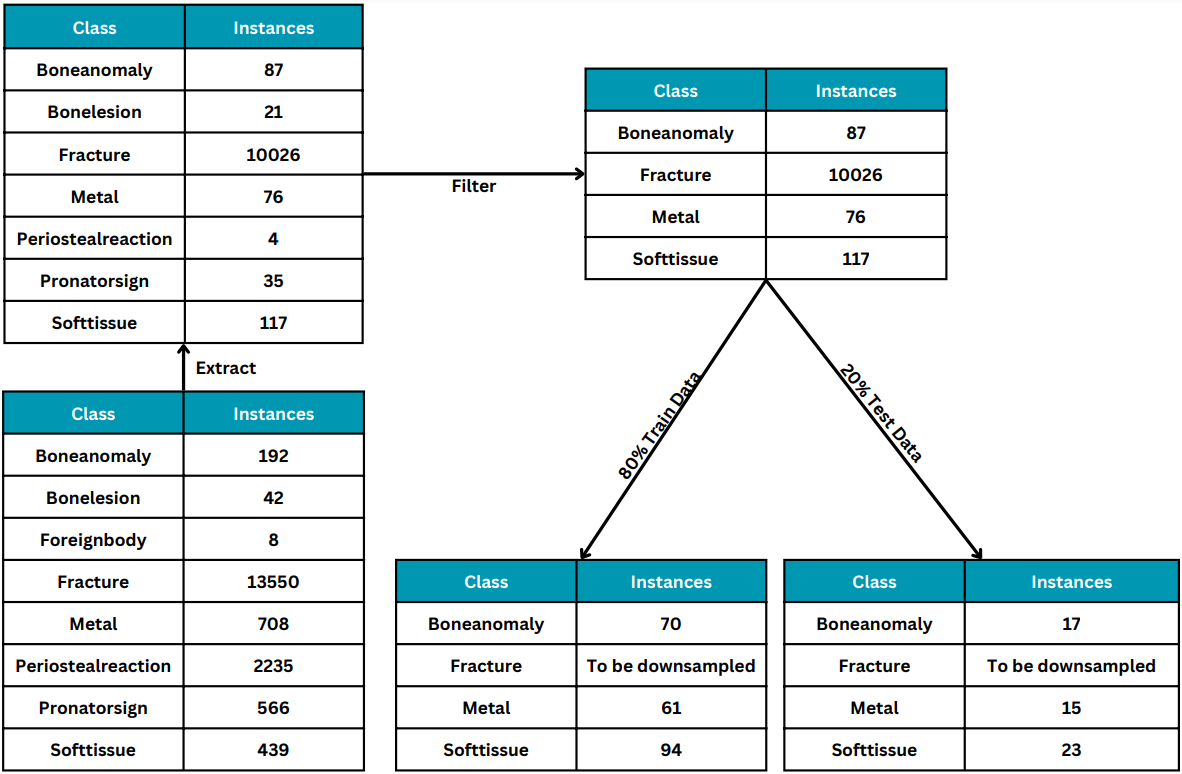}
\caption{Illustration of dataset curation steps.}
\label{fig1:dataset_curation}
\end{figure}

We address the former issue by selectively extracting images representing single classes, excluding the \enquote{foreignbody} class due to limited instances. As depicted in Fig.\ref{fig1:dataset_curation}, the first table shows the number of X-ray images associated with each pathological object; the process of image extraction significantly reduces the total number of instances. The \enquote{filter} operation eliminates classes with very few instances, which, if retained, could lead to repetitive instance representations after augmentation. This results in four final classes. For training and testing, 20\% of data from each class (except \enquote{fracture}) is allocated for testing, with the rest retained for training. We now end up with very limited data, a circumstance that mirrors the typical constraints in real-world medical studies. To address the class imbalance, we employ a downsampling strategy for the \enquote{fracture} class, and the extent of this downsampling is contingent on the number of augmentations applied to each class.

For testing purposes, we adopt two distinct approaches: in one testing set, we augment the existing images to reach a count of around 120 images per class, while in the other, we keep the original images for each class and reduce the \enquote{fracture} class to 120 and 25 images respectively. This results in two distinct testing sets: one comprising augmented images and the other retaining the original images, which we call a challenging test set, as detailed in Table \ref{tab:testsets}. 

\begin{table}[width=.9\linewidth,cols=6,pos=h]
\caption{Number of instances in two test sets.}
    \begin{tabular}{l c c}
    \hline
    Class & Test Set 1 & Test Set 2 \\
    \hline 
  Boneanomaly & 119 & 17 \\
  Fracture & 120 & 25 \\
  Metal & 120 & 15 \\
  Softtissue & 115 & 23 \\
    \hline                                                      
  \end{tabular}
  \label{tab:testsets}
\end{table}

For the training set, we adopt a strategy where we generate multiple augmented sets, each with varying numbers of downsampled \enquote{fracture} instances. To maintain class balance within each set, we apply data augmentation to the remaining instances in the classes until they match the number of downsampled \enquote{fracture} instances. As illustrated in Table \ref{tab:trainsets}, we create a total of four distinct augmented set versions. The rationale behind this approach is to systematically assess which augmented dataset yields the optimal performance, striking a balance between data quality and diversity. To evaluate the performance of the augmented sets, we utilize the augmented test set, reserving the more challenging original test set for evaluation once we have identified the most effective augmented set. We then partition the augmented sets into training and validation subsets. Specifically, 80\% of each augmented set is allocated for training, with the remaining 20\% reserved for validation, as illustrated in Table \ref{tab:split}.

\begin{table}[width=.9\linewidth,cols=6,pos=h]
\caption{Number of instances in 4 augmented sets. (The fracture class is downsampled in each case).}
    \begin{tabular}{l c c c c}
    \hline
    Class & A1 & A2 & A3 & A4 \\
    \hline 
  Boneanomaly  & 1050  & 490 & 280 & 210\\
  Fracture     & 1100  & 500 & 300 & 200\\
  Metal        & 1054  & 496 & 310 & 186 \\
  Softtissue   & 1034  & 470 & 282 & 188\\
  
  \hline
  \textbf{Total} & 4238 & 1956 & 1172 & 784 \\
    \hline                                                      
  \end{tabular}
  \label{tab:trainsets}
\end{table}

\begin{table}[width=.9\linewidth,cols=6,pos=h]
\caption{Instances from each augmented set after partitioning.}
    \begin{tabular}{c c c}
    \hline
    Augmented Set & Training Instances & Validation Instances \\
    \hline 
  A1 & 3390 & 848 \\
  A2 & 1564 & 392 \\
  A3 & 937 & 235 \\
  A4 & 626 & 158  \\
    \hline                                                      
  \end{tabular}
  \label{tab:split}
\end{table}


For the augmentation process, we employed Keras' ImageDataGenerator, utilizing the following set of parameters: 
\begin{align*}
& \textit{rotation\_range}=15, \\
& \textit{width\_shift\_range}=0.2, \\
& \textit{height\_shift\_range}=0.1, \\
& \textit{horizontal\_flip}=\textit{True}, \\
& \textit{shear\_range}=0.2, \\
& \textit{brightness\_range}=(0.7, 1.3), \\
& \textit{zoom\_range}=0.1, \\
& \textit{fill\_mode}=\textit{\enquote{constant}}, \\
& \textit{zca\_whitening}=\textit{True}.
\end{align*}
While the selection of these parameter values can be somewhat arbitrary, we opted for these specific values as they strike a balance between augmenting the images effectively without introducing excessive rotation, shift, brightness exposure, or zoom that could potentially lead to images being out of frame or overly exposed. 

\subsection{Plug-in Module For FGVR}
\begin{figure*}
\includegraphics[width=1\textwidth, height=7.5cm]{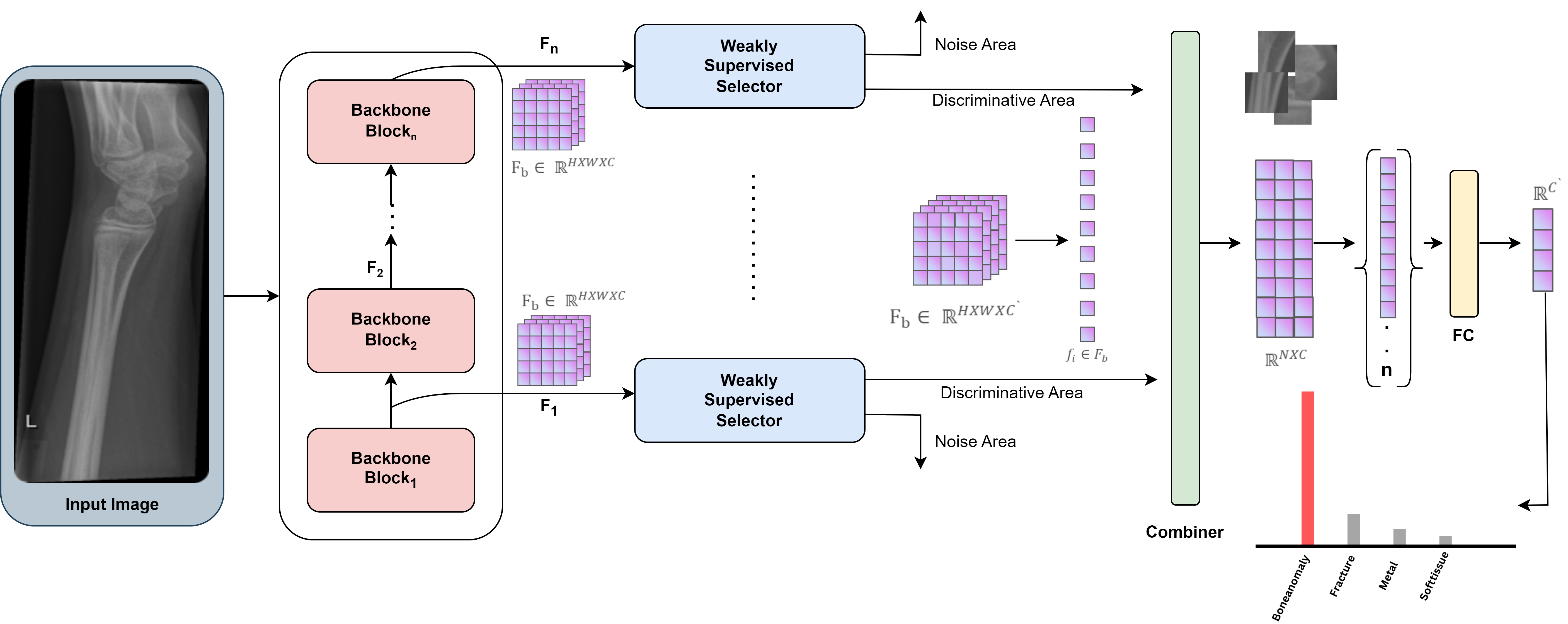}
\caption{Pipeline of plug-in module for fine-grained wrist pathology recognition. Backbone Block\(n\) denotes the \(n\textsuperscript{th}\) block within the primary network. Post image input, each block's feature map is directed to the Weakly Supervised Selector to filter discriminative regions. A Combiner then fuses these selected features to obtain prediction results.}
\label{fig1:pim_architecture}
\end{figure*}

In this study, we demonstrate the potential of fine-grained recognition in the domain of wrist pathology recognition using the recently published framework called Plug-in Module for FGVR. The network incorporates innovative background segmentation and feature combination methods, enhancing the precision of fine-grained visual recognition. We later integrate \textit{Evo\textbf{l}ved S\textbf{i}gn M\textbf{o}me\textbf{n}tum} (LION) optimizer into this network to further enhance the performance of this network. Furthermore, we carried out an ablation analysis, examining different elements of the network, such as the FPN size and selections from each backbone block. This thorough examination culminated in the recommendation of a bespoke network fine-tuned for wrist pathology recognition. We then proceed to generate heatmaps from the refined network using the explainable AI technique known as Grad-CAM \cite{selvaraju2019gradcam}. The pipeline of the network is shown in Fig.\ref{fig1:pim_architecture}.

The Plug-in Module is composed of a backbone, a selector, a combiner, and a Feature Pyramid Network (FPN). Vit \cite{Dosovitskiy2020} employs linear transformations to convert image patches into tokens, enabling global information exchange through a self-attention mechanism. However, it lacks hierarchical feature expression for local regions. In SwinTransformer \cite{liu2021swin}, the extraction of features from local regions at various scales is accomplished through a multi-layer self-attention structure. We, therefore, select SwinT as our backbone. Swin-T comprises four blocks, each generating the number of regions selected by the Weakly Supervised Selector.

The Weakly Supervised Selector operates by receiving feature maps as input. Within these maps, each feature point, represented by a pixel, is subjected to classification by a linear classifier. This process yields the identification of the most discriminative regions, while simultaneously relegating background noise to a flat probability. The selection criterion employed here is the maximum predicted probability to ascertain the significance of a feature point. This particular criterion aids in distinguishing between pivotal feature points and less relevant ones, thereby emphasizing regions with the highest likelihood of accurate classification, while minimizing the potential distraction caused by background noise. The importance of each pixel \( f_i \) in a particular feature map \( F_i \) can be determined using the softmax function as follows:
\begin{equation}
softmax(f_i) = e^{f_i} \left( \sum_{j} e^{f_j} \right)^{-1}, \quad \forall f_i \in F_i
\end{equation}
Given the probabilities of each feature point \( f_i \), we aim to select the feature points with the highest probabilities. Let \( p_i \) denote the probability of feature point \( f_i \), and assume there are \( n \) feature points in total. First, we sort the indices of the probabilities in descending order to obtain a sorted index vector \( I \) as follows:
\begin{equation} \label{eq:sort}
I = \text{argsort}^{\downarrow}(\mathbf{p}),
\end{equation}
where \( \mathbf{p} = [p_1, p_2, \ldots, p_n] \) is the vector of probabilities. Next, we select the top \( k \) indices from \( I \) to obtain a subset of indices \( K \) as follows:
\begin{equation} \label{eq:subset}
K = I[1:k].
\end{equation}
Finally, we gather the most important feature points \( f_i \) corresponding to the indices in \( K \) to obtain a set of important feature points \( \mathbf{f}_{\text{important}} \) as follows:
\begin{equation} \label{eq:gather}
\mathbf{f}_{\text{important}} = \{ f_i : i \in K \}.
\end{equation}

To effectively detect objects of varying sizes in an image, it is essential to capture features at multiple scales or resolutions. Wrist pathologies, for instance, can exhibit different sizes, making it advantageous to employ an FPN when dealing with X-ray images of fractures or other wrist pathologies. This is particularly relevant because fractures are often relatively small compared to the overall image size. FPNs facilitate the extraction of features at different levels of detail from the input image, enabling a comprehensive analysis of objects at various scales.

The employed method for feature fusion is graph convolution, wherein all selected feature points are structured into a graph. In this graph structure, nodes epitomize features situated at varying spatial locations and scales. This graph is then fed into a Graph Convolutional Network (GCN), facilitating the learning of relationships amongst different nodes. Subsequently, these feature points are consolidated into several super nodes via a pooling layer. Ultimately, the features of these super nodes are averaged, and a linear classifier is utilized to accomplish the prediction. 

Having delineated the individual components comprising the plug-in module, we shall now elucidate its operational mechanics with an illustrative example, referencing the pipeline depicted in Fig.\ref{fig1:pim_architecture}. Let the feature map output by the \(b\textsuperscript{th}\) block in the backbone network be represented as \(F_b \in \mathbb{R}^{C \times H \times W}\), where \(H\), \(W\), and \(C\) denote the height, width, and size of the feature dimension of the feature map, respectively. This feature map is then fed into a weakly supervised selector where each feature point is classified by a linear classifier, producing a new feature map \(F_b \in \mathbb{R}^{C' \times H \times W}\), where \(C'\) is the number of target classes. The class prediction probability of each feature point is obtained via the softmax function, and the feature points with the highest confidence scores are selected in the weakly supervised selector.

The selected features are fused through the fusion model. Assuming that the total number of selected feature points is \(N\), the feature maps are concatenated together along the feature dimension prior to input to the fully connected layer, as expressed below:
\begin{equation}\label{eq_concat}
    F_{\text{concat}} = \text{Concat}(f_1', f_2', \ldots, f_N') \in \mathbb{R}^{N \times C'}.
\end{equation}
Subsequently, the concatenated feature map \(F_{\text{concat}}\) is input to the fully connected layer, yielding a prediction result with dimension \(\mathbb{R}^{C'}\), as illustrated below:
\begin{equation}
    F_{\text{pred}} = F_{\text{concat}} W + b \in \mathbb{R}^{C'},
\end{equation}
where \(W\) and \(b\) denote the weight matrix and bias vector of the fully connected layer, respectively. Through this architecture, the selected local features are recombined into global features, capable of representing the entire image. It is worth noting that, to simplify the illustration, a straightforward concatenation Eq.\ref{eq_concat} is employed. However, as previously mentioned, we utilize graph convolution for fusion. This approach confers an advantage as it facilitates more efficient integration of the features at each point, without compromising the results produced by the backbone model.

This architecture aims to execute fine-grained recognition without employing any artificial labels for training besides image-level annotations. The primary training aim is to enable the features of each feature map \( F_b \) to be classified. The overall loss is computed by initially averaging the prediction outputs of all feature points as expressed in Eq.\ref{eq_avg}, where \( f_{i,s} \in \mathbb{R}^{C'} \) signifies the feature point at position \( s \) of the \(b\textsuperscript{th}\) block of the feature map, and \( S \) represents the output feature map space of this block with size \( H \times W \). Following this, the class loss for the entire block is deduced using Cross Entropy as indicated in Eq.\ref{eq_loss}
\begin{equation}\label{eq_avg}
v_i = \frac{1}{H \times W} \sum_{s \in S} f_{i,s}
\end{equation}
\begin{equation}\label{eq_loss}
Loss_b = - \sum_{i \in L} \log(v_{\textit{category},i})
\end{equation}




\subsection{{Modifications to PIM}}

{This subsection delves into the adjustments made to two fundamental components of the plug-in module: ``Number of Selections'' and ``FPN Size''. These modifications are crucial for enhancing the performance and adaptability of the architecture to our specific task of wrist pathology recognition.}

\subsubsection{Number of Selections}

{The ``Number of Selections'' parameter, denoted as} $N_{\text{sel}}$, {refers to the count of regions extracted from each backbone block by the Weakly Supervised Selector. Since our implementation utilizes the Swin Transformer (SwinT), which comprises four backbone blocks, this parameter is represented as an array of four numbers, as shown in Eq.}\ref{eq: SwinT}, {each number within this array signifies the count of selected regions for the corresponding block; for instance, the first number represents the count for the first block, and so forth.}
\begin{equation}
\label{eq: SwinT}
N_{\text{sel}} = [n_1, n_2, n_3, n_4]
\end{equation}

where $n_i$ {represents the count of selected regions for the $i$-th backbone block. We anticipate that altering the number of regions extracted will significantly impact the module's performance. Augmenting the number of extracted regions could enhance performance by furnishing the module with a richer and more varied set of information regarding the input data. Conversely, reducing the number of selected regions might prove advantageous to circumvent non-contributing areas, particularly considering the minute scale of many wrist pathologies. Upon completion of the experiments and analysis, we will interpret the results to draw conclusions regarding the optimal number of regions to extract from each backbone block for maximizing the module's performance on the dataset. }

\subsubsection{FPN Size}
{The ``FPN Size'' parameter, denoted as} $\text{FPN}_{\text{size}}$, {plays a pivotal role in determining the dimensions of the feature maps intended for processing within the Graph Convolutional Combiner, a mechanism employed for feature fusion. Let} $\text{FPN}_{\text{size}}^i$ {represent the FPN size for the} $i$-th {backbone block. The projection size} $\text{Proj}_i$, {derived from the FPN Size, is extensively utilized in orchestrating the layers of the network:}
\[
\text{Proj}_i = f(\text{FPN}_{\text{size}}^i)
\]
where $f$ {is a function that maps the FPN size to the corresponding projection size.
To investigate the impact of FPN size variations, we conduct a series of experiments where we systematically adjust the FPN size parameter and evaluate the resulting performance metrics. Specifically, we vary the FPN size within a predefined range and observe its effect on the module's performance in terms of accuracy, precision, recall, and other relevant evaluation metrics. Upon completion of the experiments and analysis, we will interpret the results to draw conclusions regarding the optimal FPN size for maximizing the module's performance on the dataset. Insights gained from this examination will inform future iterations of the module architecture and guide its refinement for enhanced performance in practical applications.}

\subsubsection{Evolved Sign Momentum (LION)}
The primary objective in deep neural networks entails the minimization of discrepancies between predicted outputs and true target values, which is gauged using a loss function. Optimizers serve as mechanisms designed to minimize this loss, thereby directing the network toward rendering more accurate predictions. Notably recognized optimizers within this realm include Adam and SGD, owing to their efficacy and operational efficiency across diverse settings. Nonetheless, Adam's memory consumption, attributed to its retention of past gradients for both momentum and RMSprop, can pose challenges, particularly with exceedingly large models or large batch sizes. Conversely, SGD's convergence trajectory tends to be more protracted relative to adaptive learning rate methods, potentially requiring a higher iteration count to reach a satisfactory solution. Moreover, SGD exhibits a pronounced sensitivity to feature scaling, often necessitating normalization to enhance performance. 

Addressing these setbacks, we employ a more resilient state-of-the-art optimizer named \textit{Evo\textbf{l}ved S\textbf{i}gn M\textbf{o}me\textbf{n}tum}, abbreviated as LION \cite{Chen2023}. This optimizer demonstrates superior memory efficiency compared to Adam, solely maintaining track of the momentum. Uniquely, unlike adaptive optimizers, its update encompasses a consistent magnitude for each parameter, derived via the sign operation. 


In our study, we also demonstrate how incorporating LION into the plug-in module for fine-grained recognition enhances its performance in comparison to utilizing SGD as its default optimizer. Algorithm \ref{Lion Optimization Algorithm} shows a simple pseudocode for LION, accompanied by an explanation of each step.

\begin{algorithm}
\caption{LION Optimization Algorithm}
\label{Lion Optimization Algorithm}
\begin{algorithmic}[1]
\State Given: $\alpha$, $\beta$, $\gamma$, $\delta$, $h$
\State Initialize: $\omega_0$, $\mu_0 \leftarrow 0$
\While{$\omega_t$ not converged}
    \State $g_t \leftarrow \nabla_\omega h(\omega_{t-1})$  
    \\
    \textbf{update model parameters}
    \State $c_t \leftarrow \alpha \mu_{t-1} + (1 - \alpha) g_t$  
    \State $\omega_t \leftarrow \omega_{t-1} - \delta_t(\text{sign}(c_t) + \gamma \omega_{t-1})$
    \\
    \textbf{update momentum} 
    \State $\mu_t \leftarrow \beta \mu_{t-1} + (1 - \beta) g_t$  
\EndWhile
\State \Return $\omega_t$
\end{algorithmic}
\end{algorithm}

Given the learning rates $\alpha$, $\beta$, regularization factor $\gamma$, update rate $\delta$, objective function $h$, the algorithm starts by initializing the model parameters $\omega_0$ and momentum $\mu_0$ to zero. The algorithm enters a loop that continues until the model parameters converge. At each step, the gradient of the objective function $h$ with respect to the parameters $\omega$ is calculated at the previous parameter values $\omega_{t-1}$. The algorithm computes $c_t$ which is a mix of the previous momentum $\mu_{t-1}$ and the current gradient $g_t$. This combination helps in smoothing the update process. The model parameters $\omega$ are updated using a unique rule that incorporates a learning rate $\delta_t$, the sign of $c_t$, and a regularization term $\gamma \omega_{t-1}$. The sign function ensures a uniform update magnitude, and the regularization term helps in controlling the parameter values. The momentum $\mu$ is updated for the next iteration, which helps in carrying forward some information from the previous gradients to smooth the update process. Finally, once the parameters have converged, the optimized parameters $\omega_t$ are returned as the output of the algorithm.

\subsection{{Grad-CAM}}
{Gradient-weighted Class Activation Mapping (Grad-CAM)} \cite{selvaraju2019gradcam} {is a method employed commonly in conjunction with convolutional neural networks (CNNs) to identify which regions of an input image are pivotal for the network’s classification decision. We have selected Grad-CAM due to its ability to be seamlessly integrated into existing neural network architectures without necessitating significant modifications or introducing substantial computational overhead. This technique utilizes the gradient information flowing into the final convolutional layer to produce class activation maps. Initially, a convolutional layer} \(B\) {is selected, which consists of multiple feature maps or "channels." Let} \(B^m\) {represent the} \(m\)-th {feature map of the chosen layer, and let} \(B^m_{xy}\) {denote the activation of the unit at position} \((x, y)\) {of the} \(m\)-th {feature map. The localization map, also referred to as a heatmap, is generated by combining the feature maps of the layer using weights} \(v^d_{m}\), {which capture the contribution of the} \(m\)-th {feature map to the network output} \(z^d\) {corresponding to class} \(d\). {To compute these weights, a specific class} \(d\) {is selected, and the degree to which the network output} \(z^d\) {depends on each unit of the} \(m\)-th {feature map is assessed by calculating the gradient} \(\frac{\partial z^d}{\partial B^m_{xy}}\) {obtained via the backpropagation algorithm. These gradients are subsequently averaged across the feature map to derive a weight} \(v^d_{m}\), {as illustrated in Equation}\ref{eq:weights}, {where} \(P\) {denotes the size (number of units) of the feature map.}
\begin{equation}
v^d_{m} = \frac{1}{P} \sum_x \sum_y \frac{\partial z^d}{\partial B^m_{xy}}
\label{eq:weights}
\end{equation}
{The subsequent step involves combining the feature maps} \(B^m\) {using the computed weights, as depicted in Equation}\ref{eq:combination}. {It is crucial to note that this combination is followed by the ReLU activation function to retain only the features with a positive influence on the class of interest. The resulting} \(M^d_{Grad-CAM}\), {referred to by the authors as the class-discriminative localization map, can be interpreted as a coarse heatmap of the same dimensions as the selected convolutional feature map.}
\begin{equation}
M_d = \text{ReLU} \left( \sum_m v^d_{m} B^m \right)
\label{eq:combination}
\end{equation}
{Once the heatmap is generated, it can be normalized and upsampled using bilinear interpolation to match the dimensions of the original image. Subsequently, it is superimposed onto the original image to highlight the regions that contribute to the network output corresponding to the selected class. This technique is highly versatile and can be applied to any differentiable network outputs.}

\subsection{Experimental Settings}
A collection of other well-established models has been employed alongside our fine-grained approach in our study. Each of these models comes with its unique architectural principles and has been benchmarked across various tasks in numerous pre-existing studies. Their inclusion in our study serves to provide a comprehensive evaluation and comparison under the defined research conditions. These models include EfficientNetV2 \cite{tan2021efficientnetv2}, EfficientNetb0 \cite{tan2020efficientnet}, NFNet \cite{brock2021highperformance}, VGG16 \cite{simonyan2015verydeep}, ViT \cite{Dosovitskiy2020}, DeiT3 \cite{touvron2022deit}, RegNet \cite{xu2021regnet}, DenseNet201 \& Dense-Net121 \cite{huang2018densely}, MobileNetV2 \cite{sandler2019mobilenetv2}, RexNet100  \cite{han2020rexnet}, ResNet50 \& ResNet101 \cite{he2015deepresidual}, ResNest101e \cite{zhang2020resnest}, InceptionV3 \& InceptionV4 \cite{szegedy2014going}, and YOLOv8 (the most recent iteration of YOLO) \cite{Ultralytics2023}. Additionally, we have employed the four most recent FGVR architectures to provide a more holistic understanding of the performance of our employed FGVR method. These are cross-layer mutual attention learning for FGVR (CMAL) \cite{LIU2023109550} FGVR via internal ensemble learning transformer (IELT) \cite{Xu2023}, {High-temperature Refinement and Background Suppression (HERBS)} \cite{chou2023finegrained}, {and MetaFormer} \cite{diao2022metaformer}.


For training, all deep neural networks maintained a consistent input resolution of \(384\) pixels necessitated by the use of the Swin-T backbone in the plug-in module. Training followed a standardized regimen with a batch size of \(16\) over \(100\) epochs. Specifically:

\begin{itemize}
  \item For the plug-in module, we initially employed the SGD optimizer with a learning rate of $\eta = 5 \times 10^{-4}$, later switching to the LION optimizer with $\eta = 5 \times 10^{-6}$.
  \item In the YOLOv8 network, the AdamW optimizer was used for the first \(10,000\) training iterations, followed by a transition to the SGD optimizer. This optimizer switching, built into the YOLOv8 codebase, aimed to optimize the training trajectory, maintaining a learning rate of $\eta = 1 \times 10^{-2}$.
  \item In contrast, the other deep neural networks in the study adhered to a consistent learning rate of $\eta = 5 \times 10^{-3}$ with the AdamW optimizer. 
\end{itemize}

\subsection{Supplementary Materials}
The supplementary material including source code, pre-trained model weights, and curated dataset split can be found at the following URLs\footnote{https://figshare.com/s/d7c612c26b8a6b51fbe1}\footnote{https://figshare.com/s/eb452d2fb36b8ae523c6}


\section{Results and Discussion}\label{sec5}
In this section, we present a comprehensive overview of the results obtained by the plug-in module for fine-grained visual recognition of wrist pathologies, as well as the performance of other state-of-the-art deep neural networks when utilized alongside the plug-in module. We initially trained the plug-in module using multiple augmented data subsets to identify the optimal performing subset, aiming to achieve an ideal equilibrium between data quality and dataset diversity. We tested the network each time on a consistent set of 474 augmented test images, distinct from the training and validation datasets. The performance of the network is shown in Table \ref{tab:table1}. The last row is for the split of the original set of images without augmentation.

\begin{table}[width=.9\linewidth,cols=6,pos=h]
\caption{Performance evaluation of the base plug-in module on a different number of augmentations of training and validation images. }
    \begin{tabular}{l c c c}
    \hline
    Set & Training & Validation & Test Accuracy\\
    \hline 
  A1 & 3390 & 848 & 83.33\% \\
  A2 & 1564 & 392 & \textbf{84.38}\% \\
  A3 & 937 & 235 & 84.18\% \\
  A4 & 626 & 158 & 78.69\% \\
  Original & 260 & 66 & 40.93\% \\
    \hline                                                      
  \end{tabular}
  \label{tab:table1}
\end{table}

Table \ref{tab:table1} highlights that utilizing 1564 training images and 392 validation images strikes an effective balance between data quality and diversity. 
Exceeding this threshold, as demonstrated in the first row, results in a decrease in network accuracy. This observation aligns with the notion that augmenting a limited set of original images excessively leads to overfitting, as the network encounters redundant image representations. Conversely, it can be seen in this case that reducing the number of images per class below the 500 threshold also leads to a decrease in accuracy. Therefore, for further analysis, we opt to adhere to the split configuration presented in the second row.

The evaluation results for each deep neural network, in conjunction with the plug-in module, are presented in Table \ref{tab:table2}. As hypothesized, it is evident that the plug-in module for fine-grained recognition surpasses all state-of-the-art neural network architectures on this specific task of wrist pathology recognition. To take it a step further and demonstrate the discriminative ability of the plug-in module on this fine-grained task, we assess its performance using the original test set, termed the challenging test set, which consists of a mere 80 images without any augmentation. From Table \ref{tab:table2}, we select the top five high-performing models for this evaluation, and the results are shown in Table \ref{tab:table3}.

\begin{table}[width=.9\linewidth,cols=4,pos=h]
\caption{Performance evaluation of different deep neural networks and plug-in module for FGVR }
    \begin{tabular}{l c c}
    \hline
    Model & Test Accuracy\\
    \hline 
  EfficientNetV2 & 53.59\% \\
  NFNet & 65.40\% \\
  VGG16 & 65.82\% \\
  ViT & 70.25\% \\
  DeiT3 & 70.89\% \\
  RegNet & 72.36\% \\
  DenseNet201 & 73.42\% \\
  MobileNetV2 & 76.37\% \\
  CMAL & 76.58\% \\
  RexNet100 & 77.43\% \\
  ResNet101 & 77.43\% \\
  IELT & 78.10\% \\
  DenseNet121 & 78.21\% \\
  ResNest101e & 78.27\% \\
  InceptionV4 & 78.69\% \\
  {MetaFormer} & {78.90\%} \\
  ResNet50 & 79.11\% \\
  InceptionV3 & 79.54\% \\
  EfficientNet\_b0 & 79.96\% \\
  YOLOv8x & 80.50\% \\
  {HERBS} & {82.70\%} \\
  \hline
  Our Approach (PIM for FGVR) & \textbf{84.38\%} \\
    \hline                                                      
  \end{tabular}
  \label{tab:table2}
\end{table}

\begin{table}[width=.9\linewidth,cols=4,pos=h]
\caption{Performance evaluation of different deep neural networks and plug-in module for FGVR on the original unaltered test set.}
    \begin{tabular}{l c c}
    \hline
    Model & Test Accuracy\\
    \hline 
  InceptionV3 & 31.25\% \\
  EfficientNet\_b0 & 31.25\% \\
  YOLOv8x & 72.50\% \\
  {HERBS} & {78.75\%} \\
  \hline
  Our Approach (PIM For FGVR) & \textbf{82.50\%} \\
    \hline                                            
  \end{tabular}
  \label{tab:table3}
\end{table}

We observe that the first CNN-based architectures (InceptionV3 and EfficientNet\_b0) yield a low accuracy of 31.25\% on this small challenging test set. Several factors could contribute to this result, including potential overfitting to the training data or the limited size of the test set. 
However, a plausible explanation in this context is that certain architectures may not be well-suited for this particular problem, given the small amount of data that we have. It's conceivable that some architectures may encounter challenges when dealing with specific data types or tasks. Making distinctions between highly looking categories might not be a task that these models are equipped with. This notion is supported by the superior performance of the plug-in module network compared to its counterparts. 
Fine-grained models often incorporate specialized techniques to focus on these small but critical differences, therefore, learning important features from even limited data. Standard CNN architectures might not be nuanced enough to capture subtle differences between very similar classes, especially without ample training data.

We further note that despite having a lower accuracy compared to the plug-in module, YOLOv8x exhibits a relatively commendable performance on this small test set. This enhanced performance, compared to the CNN-based architectures, could potentially be ascribed to the refined backbone version, the CSPDarknet53 \cite{bochkovskiy2020yolov4}, employed in YOLOv8, which aids in learning a rich feature hierarchy from the input images. Additionally, the implementation of a self-attention mechanism in the head of the YOLOv8 network enhances focus on important regions. Other elements, such as the utilization of an FPN and mosaic augmentation, may also have contributed to this performance advantage. HERBS being a fine-grained architecture, outperforms YOLOv8x by 6.25\%.

We proceed to integrate a state-of-the-art optimizer called \textit{Evo\textbf{l}ved S\textbf{i}gn M\textbf{o}me\textbf{n}tum} or LION in the plug-in module to see whether the integration of this optimizer could potentially make the plug-in module learn and generalize to the data better. The results post-integration are presented in Table \ref{tab:table4}, offering a comparison between the augmented test set consisting of 474 images and the original test set of 80 images against the default plug-in module configuration.

\begin{table}[width=.9\linewidth,cols=6,pos=h]
\caption{ Evaluation of plug-in module after integration of LION optimizer }
    \begin{tabular}{l c c}
    \hline
    Model & Test Set 1 & Test Set 2\\
    \hline 
  
  PIM  & 84.38\% & 82.50\%\\
  PIM + LION & \textbf{85.44}\% & \textbf{83.75\%}\\
    \hline               
  \end{tabular}
  \label{tab:table4}
\end{table}

The findings suggest that incorporating the LION optimizer enhances the accuracy of the plug-in module on both test sets. This is attributed to the increase in the generalization capacity of the plug-in module after the integration of LION, leading to superior performance on unseen data. This improved generalization is attributed to the sign operation in LION, discussed in section \ref{sec3}, which introduces noise to the updates, serving as a form of regularization and aiding in generalization.

\subsection{Ablation Analysis}
This subsection focuses on two integral components of the plug-in module: \enquote{Number of Selections} and \enquote{FPN Size}. 
Table \ref{tab:table5} presents an assessment of the plug-in module across varying sets of selections. It appears that the default configuration (the one that we've been using since the beginning) yields the highest accuracy, and as such, we have opted to retain the default selection of areas.

\begin{table}[width=.9\linewidth,cols=6,pos=h]
\caption{ Evaluation of plug-in module on different arrays of selections }
    \begin{tabular}{l c c }
    \hline
    Number of Selections & Test Set 1 & Test Set 2 \\
    \hline 
    
  (256,128,64,32) & 81.70\% & 81.25\%\\
  (512,256,128,64) & 83.12\% & 82.50\%\\
  (1024,512,128,64) & 83.80\% & 81.25\%\\
  (1024,512,128,128) &  84.81\% & 81.25\%\\
  (2048,512,128,128) & 82.10\% & 78.75\%\\
  (2048,512,128,32) (default) & \textbf{85.44\%} & \textbf{83.75\%}\\
  (2048,512,128,64)  & 84.60\% & 82.50\%\\
    \hline               
  \end{tabular}
  \label{tab:table5}
\end{table}

Table \ref{tab:table6} shows the assessment of the plug-in module on arbitrarily selected FPN sizes. This examination aims to discern whether an increase or decrease in FPN size augments the module's performance on this particular dataset. It is evident from the table that an FPN size of 1024 yields the highest accuracy on the augmented test set. While the default FPN size yields the highest accuracy on the original test set.

\begin{figure*}
\centering
\includegraphics[width=0.72\textwidth, height=6.8cm]{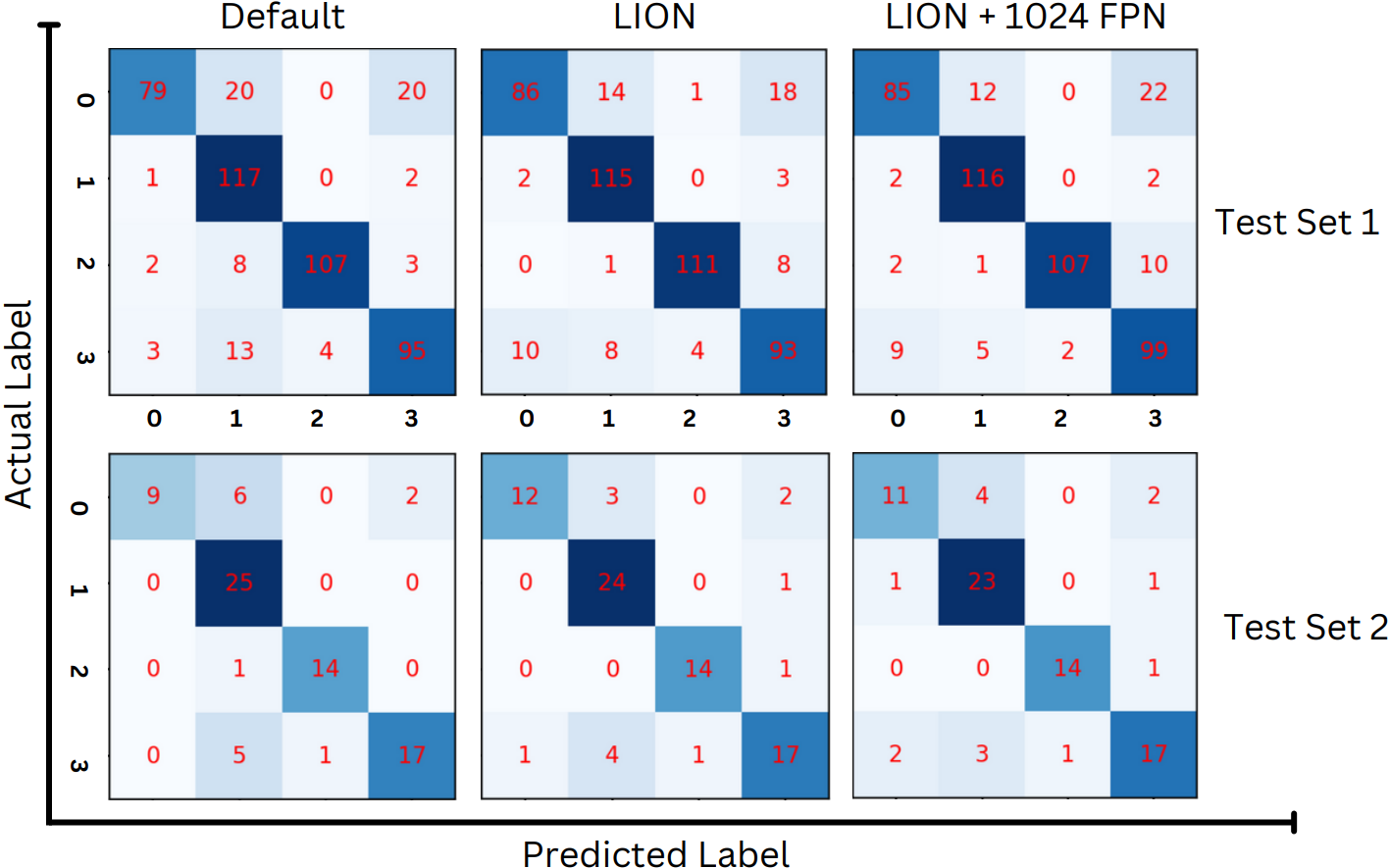}
\caption{Confusion matrices by the plug-in module across both test sets under three configurations}
\label{fig1:pim_cms}
\end{figure*}

\begin{table}[width=.9\linewidth,cols=6,pos=h]
\caption{ Evaluation of plug-in module on different sizes of FPN }
    \begin{tabular}{l c c}
    \hline
    FPN Size & Test Set 1 & Test Set 2 \\
    \hline 
    
  512 & 82.91\% & 78.75\%\\
  1024 & \textbf{85.70\%} & 81.25\%\\
  1536 (default) & 85.44\% & \textbf{83.75\%}\\
  2048 &  85.23\% & 81.25\%\\
  3000 & 81.01\% & 80.00\%\\
    \hline               
  \end{tabular}
  \label{tab:table6}
\end{table}

{Given the best configurations on the augmented test set, we execute cross-validation on the entire dataset, encompassing training, validation, and augmented testing data, totaling 2422 instances. We set} \( K = 10 \), {and for each} \( K \), {the model underwent 10 epochs. This experiment yielded a mean accuracy of} 97.55\%  {with a standard deviation of} 3.43.

Table \ref{tab:table7} encapsulates the progressive accuracy improvements we have attained with the plug-in module. A notable uptick in performance (PIM) is observed with the augmentation of training data. The subsequent integration of the LION optimizer further bolstered the performance across both sets. Ablation analysis revealed that an FPN size of 1024 enhanced the accuracy of the PIM on the augmented test set, albeit at the cost of a dip in accuracy on the original test set.

\begin{table}[width=.9\linewidth,cols=6,pos=h]
\caption{ Improvements in Accuracy of Plug-in Module for FGVR }
    \begin{tabular}{l c c}
    \hline
    Model & Test Set 1 & Test Set 2\\
    \hline 

  PIM Without Augmentation & 40.93\% & 43.75\%\\
  PIM With Augmentation  & 84.38\% & 82.50\%\\
  PIM + LION & 85.44\% & \textbf{83.75\%}\\
  PIM + LION + 1024 FPN & \textbf{85.70}\% & 81.25\%\\
    \hline               
  \end{tabular}
  \label{tab:table7}
\end{table}

\subsection {Sensitivity, Specificity, and Precision Analysis}
Having completed the accuracy analysis, we now shift our focus to more pertinent metrics including precision, specificity, and sensitivity for the plug-in module across both test sets. The evaluation metrics take on nuanced importance to ensure precise and reliable model performance and although our dataset is balanced, which is advantageous as it reduces the bias towards any particular class, the criticality of each metric remains significant. Sensitivity is paramount in ensuring that all actual cases of each pathology are correctly identified. For instance, missing a fracture or a metal object could have severe implications for the patient. High sensitivity ensures that the model captures most of the true positive cases, thus potentially aiding in timely and accurate diagnosis and treatment. Specificity, or the ability to correctly identify negative cases for each class, is essential to reduce false alarms and ensure that the true nature of the pathology is captured. For example, a high specificity would mean that individuals without a fracture are correctly identified as such, reducing the chances of unnecessary further examinations or treatments. High precision ensures that when the model predicts a particular type of pathology, it is indeed correct a high proportion of the time. For instance, incorrectly identifying a fracture as a bone anomaly might lead to inappropriate or delayed treatment which can worsen the patient's condition. These metrics ensure that the model not only accurately identifies the type of pathology but also minimizes the chances of misdiagnosis, which is crucial in medical applications for ensuring patient safety and effective treatment planning.

We initiate our examination by reviewing the confusion matrices produced by the plug-in module across both test sets under three distinct configurations: employing the default settings, post-integration of the LION optimizer, and adjusting the FPN size to 1024. The depicted matrices can be referenced in Fig.\ref{fig1:pim_cms}. Within these matrices, the first class corresponds to bone anomaly denoted as \enquote{0}, the second to fracture denoted as \enquote{1}, the third to metal denoted as \enquote{2}, and the fourth to soft tissue denoted as \enquote{3}. Observing the first row of evaluation on test set 1, it's evident that the default configuration is leading to numerous misclassifications concerning the bone anomaly class, often misidentifying it as either fracture or soft tissue class, consequently diminishing the sensitivity of the bone anomaly class. This issue appears to be mitigated by post-LION integration as well as upon adjusting the FPN size to 1024. The combination of LION integration and a 1024 FPN size seems to enhance performance notably for the soft tissue class, whereas employing LION alone yields better performance for both bone anomaly and metal classes. When examining test set 2, the LION optimizer once again demonstrates a superior performance in accurately classifying the bone anomaly class. 

Table \ref{tab:table8} presents the evaluation of the plug-in module under the employed configurations, utilizing metrics such as sensitivity, specificity, and precision on test set 1. Table \ref{tab:table9} presents the same configuration but for test set 2. For test set 1, the following are the key highlights: Class 0 exhibited a significant improvement in sensitivity, increasing from 66\% to 72\% after LION integration. However, this enhancement came with a minor decline in precision. Additionally, when the FPN was adjusted to 1024, precision also decreased for Class 0. The LION optimizer can influence various model hyperparameters, and alterations in one component may require adjustments in other hyperparameters to uphold the desired balance between precision and recall for both classes of bone anomaly and soft tissue. Furthermore, the optimizer's use of a sign function for uniform updates can lead to more substantial and consistent parameter adjustments during iterations, promoting faster convergence. However, this rapid convergence may prioritize sensitivity (recall) over precision for some classes by limiting the model's ability to refine class boundaries thoroughly. In medical contexts, sensitivity holds paramount importance as a metric. Hence, prioritizing an increase in sensitivity, even if it leads to a slight reduction in precision, remains our focus. Class 1 maintained consistent sensitivity across configurations, peaking at 97\% with both the default setup and following FPN size adjustment. Class 2's sensitivity rose from 89\% to 93\% by post-LION integration. The highest sensitivity for Class 3, at 86\%, was achieved after adjusting the FPN size. {The peak average specificity was 96\%, achieved by post-LION integration, while the LION with 1024 FPN size yielded the highest AP of 87\% as well as the highest F1 score of 86\%.}

Turning to test set 2, LION's integration markedly boosted the sensitivity for class 0 from 53\% to 71\%. However, there was a minor decline for class 1, dropping from a perfect 100\% to 96\%. Sensitivities for other classes remained consistent. {The peak average specificity and F1 score were 94\% and 84\%, respectively, achieved by post-LION integration, while the default configuration yielded the highest average precision of 88\%.}

{The observed performance differences across the four distinct classes merit a detailed examination to understand the underlying factors contributing to this variability. Notably, the configurations demonstrate robust performance for certain classes, such as fractures and metal anomalies, while exhibiting comparatively poorer performance for bone anomalies and soft tissue pathologies. Fracture detection benefits from discernible features such as small cracks in the bone structure, facilitating accurate classification by the algorithm. Similarly, the distinct characteristics of metal anomalies are often identifiable, allowing the fine-grained algorithm to effectively classify them. Soft tissue pathologies present a moderate challenge, yet they are typically distinguishable due to observable swollen tissue surrounding the bone, enabling accurate classification. In contrast, bone anomalies pose the greatest challenge due to their direct relation to the intricate structure of the bone itself. The algorithm encounters difficulty in accurately delineating bone structures, contributing to lower performance in this class. Instances of misclassification, such as bone anomalies being erroneously classified as fractures or soft tissue pathologies, further underscore the complexity of accurately detecting these anomalies.}

\begin{figure*}
\centering
\includegraphics[width=0.65\textwidth, height=8cm]{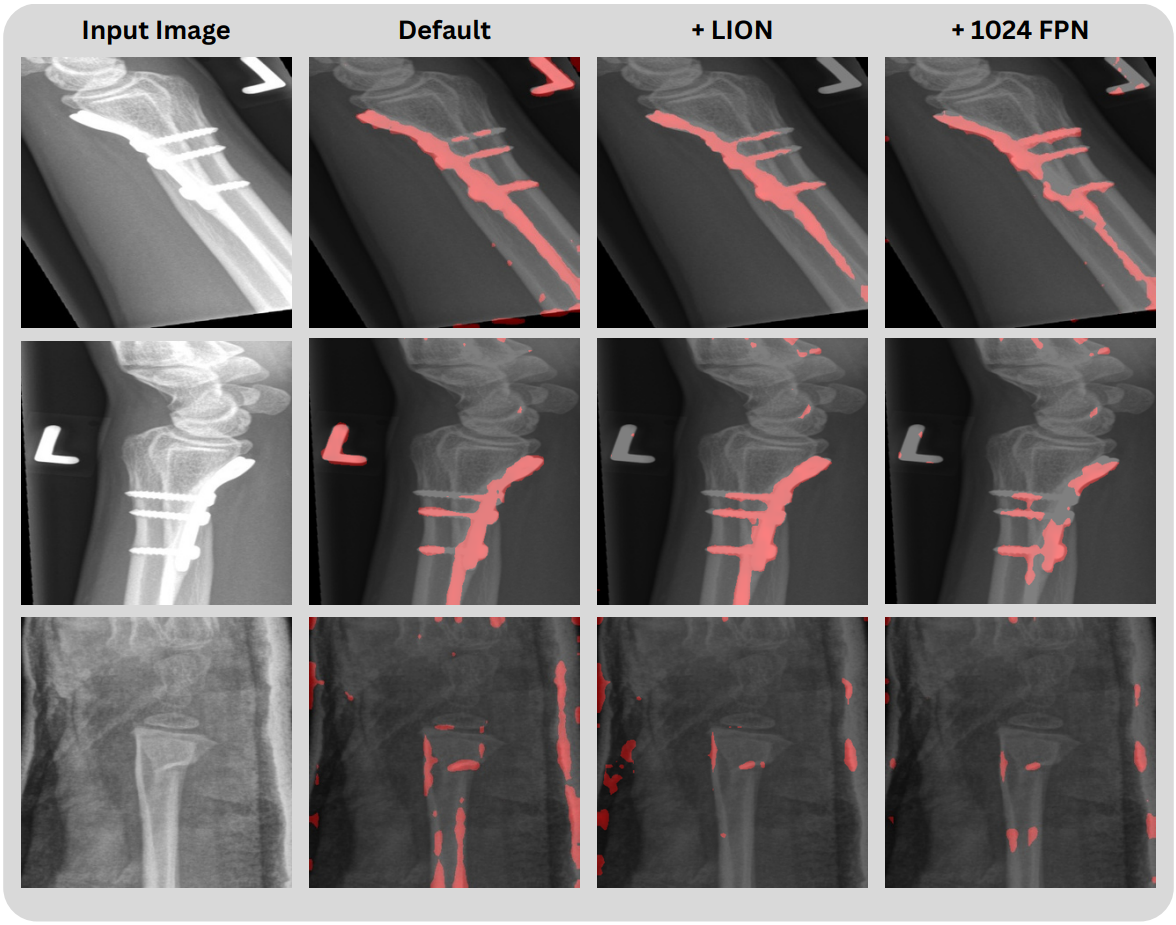}
\caption{Heatmaps produced from the three employed configurations of the plug-in module. The top two rows display heatmaps for sample images from the metal class, while the third row showcases those from the fracture class.}
\label{heatmaps}
\end{figure*}

\begin{table}[h]
    \centering
    \caption{Evaluation of plug-in module using different configurations using sensitivity, specificity, and precision (Prec.), and F1 score on test set 1.}
    \begin{tabular}{p{1.2cm}ccccc}
        \hline
        Config & Class & Sensitivity & Specificity & Prec. & {F1}\\
        \hline
        \multirow{4}{*}{PIM} & 0 & 66\% & 98\% & 93\% & {77\%}\\
        & 1 & 97\% & 88\% & 74\% & {84\%}\\
        & 2 & 89\% & 99\% & 96\% & {92\%}\\
        & 3 & 83\% & 93\% & 79\% & {81\%}\\
        \hline
        \multirow{4}{*}{+LION} & 0 & 72\% & 97\% & 88\% & {79\%}\\

        & 1 & 96\% & 94\% & 83\%  & {89\%}\\
        & 2 & 93\% & 99\% & 96\% & {94\%}\\
        & 3 & 81\% & 92\% & 76\% & {78\%}\\
        \hline
        \multirow{4}{*}{+1024FPN} & 0 & 71\% & 96\% & 87\% & {78\%}\\
        & 1 &  97\% & 95\% & 87\% & {92\%}\\
        & 2 & 89\% & 99\% & 98\% & {93\%}\\
        & 3 & 86\% & 91\% & 74\% & {80\%}\\
        \hline
    \end{tabular}
    \label{tab:table8}
\end{table}

\begin{table}[h]
    \centering
    \caption{Evaluation of plug-in module using different configurations using sensitivity, specificity, and precision (Prec.), and F1 score on test set 2.}
    \begin{tabular}{p{1.2cm}ccccc}
        \hline
        Config & Class & Sensitivity & Specificity & Prec. & {F1}\\
        \hline
        \multirow{4}{*}{PIM} & 0 & 53\% & 100\% & 100\% & {69\%}\\
        & 1 & 100\% & 78\% & 68\% & {81\%}\\
        & 2 & 93\% & 98\% & 93\% & {93\%}\\
        & 3 & 74\% & 96\% & 89\% & {81\%}\\
        \hline
        \multirow{4}{*}{+LION} & 0 & 71\% & 98\% & 92\% & {80\%}\\
        & 1 & 96\% & 87\% & 77\% & {85\%}\\
        & 2 & 93\% & 98\% & 93\% & {93\%}\\
        & 3 & 74\% & 93\% & 81\% & {77\%}\\
        \hline
        \multirow{4}{*}{+1024FPN} & 0 & 65\% & 95\% & 79\% & {71\%}\\
        & 1 &  92\% & 87\% & 77\% & {84\%}\\
        & 2 & 93\% & 98\% & 93\% & {93\%}\\
        & 3 & 74\% & 93\% & 81\% & {77\%}\\
        \hline
    \end{tabular}
    \label{tab:table9}
\end{table}

\subsection{Results Interpretability with XAI}
We generated heatmaps using the plug-in module trained on a limited dataset, attempting to visualize the discriminative regions based on which the classifications are being made. To do so, we utilize an explainable AI technique known as Grad-CAM. Fig.\ref{heatmaps} shows heatmaps derived using the three employed configurations of the plug-in module. The upper two rows illustrate heatmaps for representative images of the metal class, whereas the last row does so for the fracture class.

From the sample image representing the metal class in the first two rows, it is evident that despite the over-exposure of the image, the plug-in module successfully identified the class and emphasized the metal presence. Notably, in numerous cases, radiologists would deem such an image undiagnosable by conventional standards \cite{Atkinson2019}. It is also evident from the first two rows that the default configuration, although highlighting the metal, also emphasizes non-discriminative regions, such as the letter \enquote{L} indicating the side of the arm. {This could stem from the default configuration's potential overfitting to the presence of the letter "L" in the training data, erroneously associating it with discriminative regions and consequently highlighting irrelevant areas. The concurrent highlighting of non-relevant features alongside discriminative areas complicates the interpretability of the algorithm's output, making it challenging to distinguish between pathology and algorithmic artifacts. However, the integration of LION in the first row appears to alleviate this issue, suggesting that LION enables the algorithm to generalize effectively and avoid overfitting to non-discriminative regions.} In the second row, the FPN adjustment seems to have left out some of the discriminative regions. In contrast, the configuration after LION integration exclusively highlights the metal. Similarly, superfluous regions appear to be filtered out for the last row after integrating LION and adjusting the FPN size to 1024. 


{We conclude this section by mentioning some limitations inherent in our approach. While the quality of the heatmaps remains relatively high considering the constraints of the limited dataset, there is evidence of potential overfitting to specific regions, notably observed in Fig sure }\ref{heatmaps}, {such as the prominence of the letter ``L'' by the default configuration. Moreover, our current approach encountered challenges in accurately classifying instances belonging to the "boneanomaly" class, primarily due to their close association with the intricate structure of the bone itself. The algorithm struggled to delineate bone structures accurately, leading to decreased performance in this class and instances of misclassification, where bone anomalies were incorrectly classified as fractures or soft tissue pathologies. The heatmaps also show some difficulty in pinpointing the fractures before the integration of LION. Nonetheless, we posit that the integration of LION and a larger dataset holds promise in mitigating these challenges.}

\section{Conclusion and Future Work}\label{sec6}
The present study demonstrates that when we approach wrist pathology recognition in terms of a fine-grained recognition problem, the results outperform many of the contemporary recognition techniques on a very limited dataset. We validated this using the recent plug-in module framework designed for fine-grained recognition. We further refined the network’s performance by integrating the LION optimizer, observing a notable improvement in its generalization capability and overall performance. Lastly, we utilized the explainable AI technique called Grad-CAM to identify discriminative regions within the X-rays. Despite training our network on a constrained dataset, we found that the heatmaps demonstrated relatively high quality. We posit that utilizing larger datasets with only image-level annotations could further enhance the heatmap quality. 

Looking ahead, there are several promising directions for further research. One avenue involves refining fine-grained architectures tailored specifically for wrist pathology recognition. While current networks address general fine-grained problems, we envision future advancements in our approach could potentially eliminate the need for manual annotation entirely, thereby reducing the future costs of medical data acquisition. Additionally, we see the potential for improvement through the utilization of larger datasets for training fine-grained networks, enhancing the quality of heatmaps. {Moreover, although we performed a cross validation test on the entire dataset, the absence of an external validation set represents a limitation. This limitation may compromise the model's ability to generalize to unseen data, particularly when confronted with variations in imaging techniques, patient demographics, and clinical practices across different healthcare settings. Pediatric wrist pathologies may exhibit variability in prevalence, severity, and diagnostic patterns across different regions and demographic groups. Without external validation, it becomes challenging to evaluate the model's performance robustness and its potential biases towards specific patient subgroups or disease presentations. Future research could address this by conducting experiments using our approach on validation sets.} Lastly, exploring the benefits of applying the LION optimizer on larger wrist X-ray datasets is of particular interest, given its proven performance enhancements with more extensive training sets. It is advisable to further optimize specific hyperparameters within LION to enhance metrics beyond sensitivity, particularly precision. Notably, we observed a decline in precision alongside the increase in sensitivity for some classes, indicating a need for fine-tuning to achieve a more balanced enhancement across multiple metrics.




\printcredits
\bibliographystyle{cas-model2-names}

\bibliography{Bibliography}

\end{document}